\def\year{2019}\relax
\documentclass[letterpaper]{article} 
\usepackage{aaai19}  
\usepackage{times}  
\usepackage{helvet}  
\usepackage{courier}  
\usepackage{url}  
\usepackage{graphicx}  
\usepackage{amsfonts}  %
\usepackage[top=2cm, bottom=2cm, left=2cm, right=2cm]{geometry}
\usepackage{algorithm}
\usepackage{algorithmicx}
\usepackage{algpseudocode}
\usepackage{amsmath}
\begin{document}
%
\title{Solve Traveling Salesman Problem by Monte Carlo Tree Search and Deep Neural Network}
\author{
Zhihao Xing,\textsuperscript{\rm 1} Shikui Tu,\textsuperscript{\rm 1} Lei Xu\textsuperscript{\rm 1}. \\
Department of Computer Science and Engineering, Shanghai Jiao Tong University, Shanghai, China
}
\maketitle
\begin{abstract}
	We present a self-learning approach that combines deep reinforcement learning and Monte Carlo tree search to solve the travelling salesman problem. The proposed approach has two advantages. First, it adopts deep reinforcement learning to compute the value functions for decision, which removes the need of hand-crafted features and labelled data. Second, it uses Monte Carlo tree search to select the best policy by comparing different value functions, which increases its generalization ability. Experimental results show that the proposed method performs favorably against other methods in small-to-medium problem settings. And it shows comparable performance as state-of-the-art in large problem setting.
\end{abstract}

\section{Introduction}
Travelling salesman problem(TSP) enjoys a long history and has many practical applications in real life. Its goal is to find the shortest route that visits each city once and ends in the origin city. Despite the importance of the problem, it is well-known as a NP-hard problem\cite{papadimitriou1977euclidean}. 
%

Traditional methods for solving TSP can be categorized into three directions. First, all permutations are traversed to search for the optimal solution, which is only limited to small-scale problem. Second, approximation algorithms are applied to solve the problem, but the best solution cannot be guaranteed. Third, heuristic algorithms can be used to find a satisfactory solution within a reasonable time, but it requires well-designed heuristics to assists in the search.

Recent advances in deep learning have achieved an amazing breakthrough in many fields \cite{Krizhevsky2012ImageNet,Graves2013Speech}. Most of these achievements benefit from supervised learning where various neural network architectures are proposed, including multi-layer perceptrons \cite{Rosenblatt1960Perceptrons}, convolutional networks \cite{Lecun1989Backpropagation} and so on. However, training a deep neural network requires a huge number of data. For example, the most famous dataset \cite{Deng2009ImageNet} for image classification has about 3.2 million images. But for TSP, we cannot easily obtain so much ground truth data. Therefore, researches have adopted reinforcement learning to allow the network to learn by rewards and punishments. 


Monte Carlo tree search (MCTS) has become a popular approach to solve two-player game problems since the appearance of AlphaGo Zero \cite{Silver2016Mastering}. With the help of deep neural network, MCTS can solve problems with a tremendously large solution space. Researches have applied MCTS to find solutions for other problems similar to TSP\cite{rimmel2011optimization,bnaya2011repeated}.

In this paper, we present a new self-learning approach with the combination of deep reinforcement learning and Monte Carlo tree search to solve the famous travelling salesman problem. On 2D Euclidean graphs with up to 100 nodes, the proposed method significantly outperforms the supervised-learning approach \cite{Vinyals2015Pointer} and obtains performance close to reinforcement learning approach \cite{Dai2017Learning}. 

The remainder of the paper is organized as follows: After related work reviewed in Section 2, we introduce the proposed DMR framework in Section 3. Experimental results are shown in Section 4. In Section 5, we come to our conclusion and future work.

\section{Related Work}
In this section, we introduce three different directions to solve the TSP problem.

\subsection{Shallow Neural Network}
In 1985, Hopfield \emph{et al.} proposed a neural network to solve TSP\cite{hopfield1985neural}. This is the first time that researchers attempted to use the neural network to solve combinatorial optimization problems. Since the impressive results produced by this approach, many researchers have made efforts on improving the performance\cite{den1988traveling,alan1988alternative}. Many shallow network architectures were also proposed to solve the combinatorial optimization problem\cite{favata1991study,fort1988solving,angeniol1988self,kohonen1982self}.

\subsection{Deep Neural Network}
Recently years, deep neural networks have been adopted to solve TSP. Vinyals \emph{et al.} introduced a neural architecture called Pointer Network(Ptr-Net)\cite{Vinyals2015Pointer}. Ptr-Net is a simple model based on sequence-to-sequence model. Compared to the sequence-to-sequence model, Ptr-Net introduces an attention mechanism to output a dictionary whose length is proportional to the input sequence. Two flaws exist in the network. First, Ptr-Net can only be applied to solve problems of small scale. If the number of cities reaches 40, the performance of the algorithm suffers greatly. Second, invalid routes might be generated by the approach. For example, it might output a route with two repeated cities.

\subsection{Deep Reinforcement Learning}
With the use of deep reinforcement learning, deep Q-networkn\cite{Mnih2013Playing} becomes a general framework that is applied in many different methods including \cite{Bello2016Neural,Dai2017Learning}.

Bello \emph{et al.} proposed Neural Combinatorial Network\cite{Bello2016Neural} to combine neural network and reinforce learning to deal with combinatorial optimization problem. This framework consists of two stages, the \textit{RL pretrained} stage and the \textit{active search} stage. The first stage is responsible for optimizing a recurrent neural network, and the second stage is to iteratively optimize the RNN with the expected reward objective.

Dai \emph{et al.} proposed a method called S2V-DQN \cite{Dai2017Learning} which combines graph embedding  and reinforcement learning. The method can extract topological information between different nodes in a graph. As a result, the approach can generalize to large-scale graphs even trained on small-scale instances with the help of graph embedding.

Kool \emph{et al.} also combined deep neural network and reinforcement Learning to solve TSP\cite {Kool2018Attention}. They integrate Attention Mechanism \cite{vaswani2017attention} into their framework, where encoder and decoder are both entirely based on attention. They improved the state-of-art performance among 20, 50 and 100 cities. However, the pretrained network has to precisely match the problem scale, which weakened the generalization ability of their framework.

\section{Proposed Approach}
This section describes our novel approach to solving combinatorial optimization problems, which, as shown in Fig.xx, consists of three modules: deep neural network, Monte Carlo tree search, and reinforcement learning. In our framework, the original problem that finding an optimal solution in the graph is converted into searching the least-cost path in a tree. The deep neural network is responsible for extracting topological information as node's features from the graph as an alternative to designing features manually. The Monte Carlo tree search is used to narrow the search space with the help of value function module of deep neural networks. We follow reinforcement learning paradigm to generate experience used to train deep neural network. We empirically demonstrate that our approach can start from initial random choice to converge to the optimal solution.

Problem-solving tasks are typically implemented in a large number of steps. At each step, there are a number of branches among which one is selected to be implemented. The traveling salesman problem can also be solved according to the above process.
\\
We use a node $v_i \in {R}^2$ to represent a city. Then one instance of the TSP problem can be described by a undirected weighted graph $G(V,E,W)$, where $V$ is the set of finite nodes, $e_{ij}=(i,j)\in E$ is the edge between $v_i$ and $v_j$, and $w_{i,j} \in W$ is the weight of edge $e_{ij}$ $\left(w_{ij}=w(e_{ij}), w:E \rightarrow R^{+}\right)$. Given a set of cities, we are concerned with finding the path traversing each city once, which is noted as a tour, and has the shortest length.

We convert the original problem of finding the shortest tour in a graph to searching a path with the least cost in the tree.

\subsection{Tree Search}
Tree search methods aim to find the optimal path in a tree.
We use $S=\{v_1,v_2,...,v_i\}$ represent a path started with $v_1$ and ended with $v_i$, so $S$ is an ordered sequence of traversed cities. We use $\bar{S}=\{v_1,v_2,...,v_j\}$, $v_j \notin S$ denotes the set of non-traversed cities.

In tree search, the traversed path $S=\{v_1,v_2,...,v_i\}$ denotes one $state$ where $S=\{v_{start}\}$ denotes the root of the tree and the leaf node corresponds to $S=V$. Tree search needs to select the best node in the candidate sets $\bar{S}$  step by step according to the present state. There are two traditional methods called Breadth-First-Search (BFS) and Depth-First-Search (DFS), but both of them have the complexity of the order $O(b^d)$ in not only the worst sense but also the average sense.

Monte Carlo tree search \cite{Pearl1984Heuristics,Kocsis2006Bandit} is a heuristic search algorithm for some kinds of the decision process, most notably those employed in gameplay such as Total War and Go game. Different from DFS and BFS, Monte Carlo tree search aims to get the most promising moves and consists of the random sampling of the search space in tree search. Before making a decision, MCTS repeats the process called $playouts$ for many times and at each time $playout$ consists of four steps, which is illustrated in Figure 2.\\ 
\textit{\textbf{Selection}}: Start from root node $R$ and then select a child node of R according to a default policy. The newly selected node will be the root node and then repeat the above process until a leaf node $L$ is reached.\\
\textit{\textbf{Expansion}}:Create one or more child nodes of $L$ and select one node $N$ unless the game ends.\\
\textit{\textbf{Simulation}}: Start with node $N$ and play with a random strategy such as uniform random move until the game is over.\\
\textit{\textbf{Backpropagation}}: Update node information on the path from node $N$ to node $R$ using the result of the random game.\\

\begin{figure}[h]
	\includegraphics[width=0.45\textwidth,height=0.35\textheight]{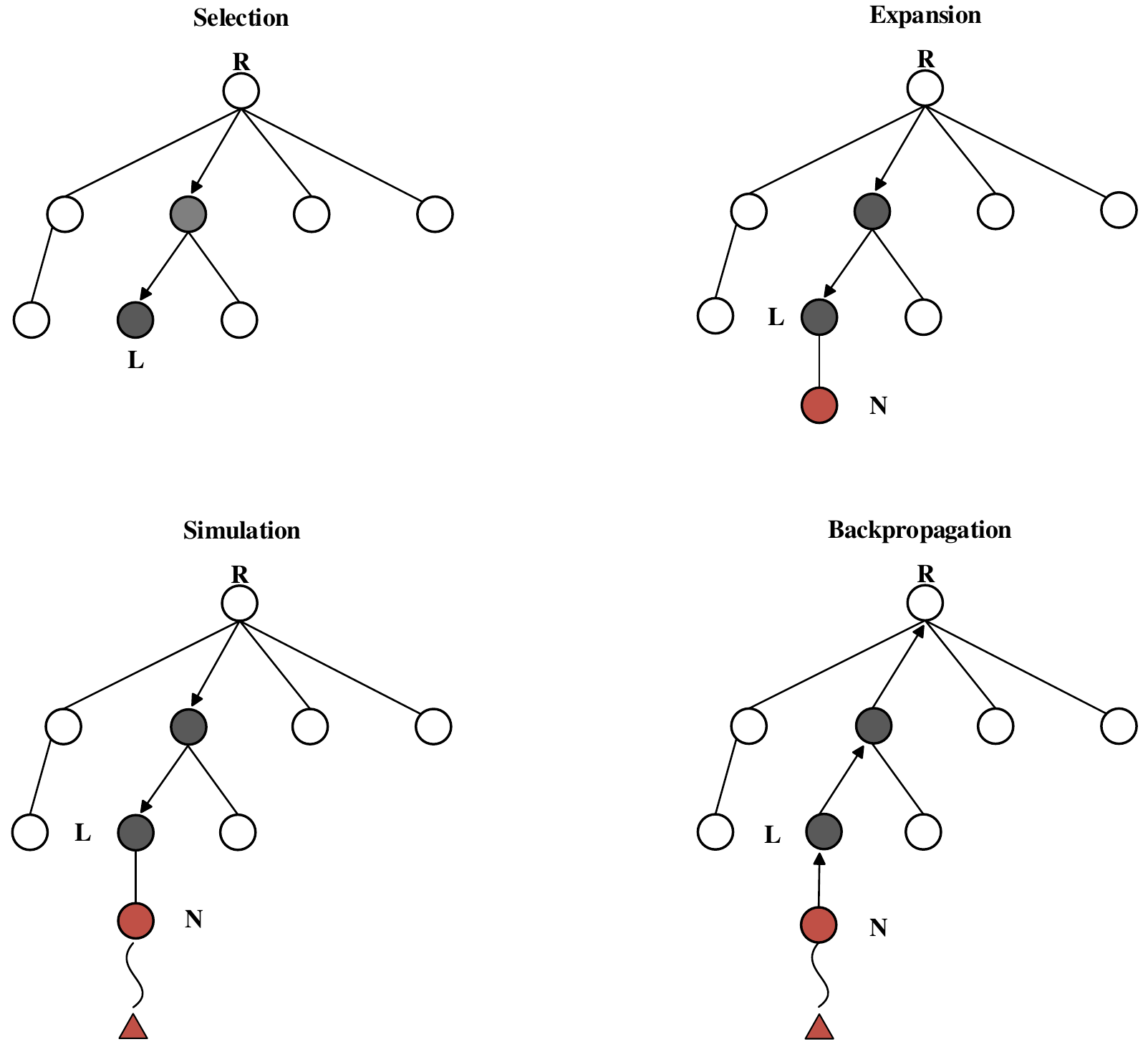}
	\caption{Monte Carlo tree search
	}
	\label{mcts}
	
\end{figure}

For traveling salesman problem, we propose an adapted version of MCTS. The details of the four phases of MCTS is as follows:\\
\textit{\textbf{Selection Strategy}.}  \cite{Kocsis2006Bandit} proposed one selection strategy called Upper Confidence bounds applied to Trees (UCT), which has achieved great success in the game. There are some differences between game-playing and combinatorial optimization problems. Firstly, a branch with the highest average rate of winning is preferred in game-playing while combinatorial optimization aims to find the extreme, which may locate in the direction without a good average value.
So given a node $s$, we modify the policy of UCT to selecting child $i$ of $s$ that  maximizes the following formulation,
\begin{equation}\label{formula_4}
\mathop{\arg\max_{\hat{Q}_{i}} \left(\ \hat{Q}_i +  C_p \sqrt{\frac{\ln N_{s}}{N_i}}\right)}
\end{equation}
where $\hat{Q}_{i} =-f(i)$ , $f$ is defined as follows,
\begin{equation}\label{formula_1}
f(v) = g(v) + h(v)
\end{equation}
where $g(v)$ is known and represents the actual length of ordered sequence $S$ from the first node to the last node, $h(v)$ is unknown and supposed to be the optimal length from $v$ to the goal $G$. In our framework, $h(v)$ is evaluated by a deep neural network, which will be described in the next section. $\hat{Q}_i$ is the best reward found under subtree of node $i$ . $N_s$ and $N_{i}$ are the number of visits of node $s$ and node $i$ respectively. $C_p > 0$ is a parameter used to balance exploitation and exploration. 

What's more, the range of $\hat{Q}$ value is different between game-playing and combinatorial optimization problems. In game-playing, the result of a game is composed of $loss$, $draw$, and $win$, i.e., $\{0,0.5,1\}$. The average reward of a node always stays within $[0,1]$. In the combinatorial optimization problems, an arbitrary returned reward may not fall in the predefined interval. Thus, we normalize the best reward of each node $c$ whose parent is node $p$ to [0,1] with the following formulation,
\begin{equation}\label{formula_5}
\hat{Q}_{c} = \frac{\hat{Q}_{c}-\hat{Q}_{min}}{\hat{Q}_{max}-\hat{Q}_{min}}
\end{equation}
where $\hat{Q}_{max}$ and $\hat{Q}_{min}$ are the maximum and minimum reward among all children nodes of node $p$ respectively.\\
\noindent \textit{\textbf{Expansion Strategy}.} When a leaf node $l$ is reached, we expand the node until its visitation count reaches a preset threshold(we set this threshold to 40). This avoids generating too many branches so as to distract the search and save computation resource. Similar to A* algorithm \cite{hart1968formal}, we expand all children nodes of the leaf node $l$ at the same time.\\
\noindent \textit{\textbf{Simulation Strategy}.} We use value function $h$ in Equation \ref{formula_3} to evaluate all children nodes which are expanded in the expansion stage.\\
\noindent \textit{\textbf{Back-Propagation Strategy}.} Instead of propagating a child node's simulation reward, we choose to use the best reward among all children nodes to back propagate to the root.

\subsection{Neural Network Architecture}
Inspired by graph embedding network \cite{dai2016discriminative,Dai2017Learning}, we propose to use graph convolutions to extract features from the graph. Each node in the graph is represented by a feature vector and merges its neighbor nodes' information recursively according to the graph topology. For each node, the feature is expressed as a 9-dimensional vector. We use an element 0 or 1 to represent whether one node has been traversed or not. Besides current node information (traversal state, x-coordinate, y-coordinate), we especially take notice of the first and the last node in the traversed path due to the solution path is the Hamiltonian path. What's more, we use edge weight as supplementary feature.

We now describe the parameterization of graph convolutions using the graph embedding. We map the features of each node $v$ in the graph to the hidden space by using the following formula:
\begin{equation}\label{formula_2}
H^{(t+1)}_{v} = \sigma (\theta_{1}x_v+\theta_{2}\sum_{u \in \mathcal{N}(v)}H^{(t)}_u+\theta_{3}\sum_{u \in \mathcal{N}(v) }\sigma(\theta_{4}w_{v,u}))
\end{equation}
where $\theta_{1} \in \mathbb{R}^l$, $\theta_{2},\theta_{3} \in \mathbb{R}^{l \times l}$ and $\theta_{4} \in \mathbb{R}^l$ are the parameters, and $\sigma$ is the rectified linear unit (relu). $x_v$ and $w_{v,u}$ are the node's features and distance \footnote{Euclidean distance: given two points $(x_1,y_1)$ and $(y_1,y_2)$ in two-dimensional plane, $D= \sqrt{(x_2-x_1)^2+(y_2-y_1)^2}$} between two nodes mentioned above respectively. And $\mathcal{N}(v)$ denotes the neighbor nodes of node $v$.

\begin{figure*}[h]
	\includegraphics[width=0.9\textwidth]{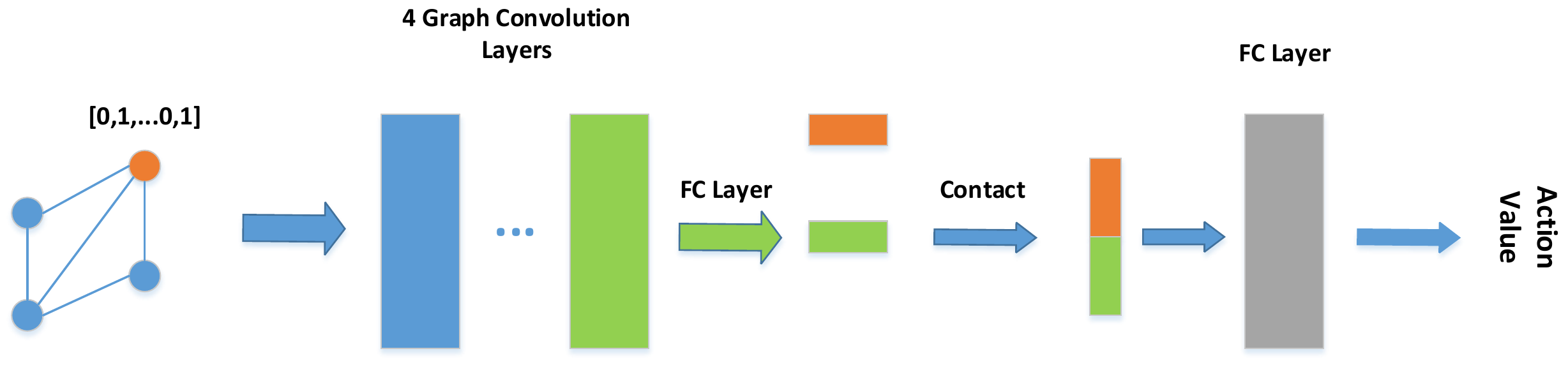}
	\centering
	\caption{Neural network architecture. Each node in the graph is embedded to $l$-dimensional vector after the Graph Convolution Layers. The first fully connected layer (green) is responsible for integrating all nodes' embedded features in the graph. The last  fully connected layer (gray) predicts the value of the selected node (orange).
	}
	\label{neural network}
	
\end{figure*}

After T iterations each node is embedded in the graph, we will use these embedding information to define $h(v)$ mentioned in Equation (\ref{formula_1}). Similar to \cite{Dai2017Learning}, we compute $h(v)$ as follows,\\
\begin{equation}\label{formula_3}
h(v) = \theta_{5}^{\top} \sigma ([\theta_{6} \sum_{u \in V} H^{(T)}_{u}, \theta_{7} H^{(T)}_{v}])
\end{equation}
where $\theta_{5} \in \mathbb{R}^{2p}$, $\theta_{6},\theta_{7} \in \mathbb{R}^{p \times p}$ and $[
\cdot,\cdot]$ denotes the concatenation operator. As suggested by \cite{dai2016discriminative}, the number of iterations T for graph embedding is 4. The architecture of the neural network is illustrated in Figure \ref{neural network}.

\textbf{}

\subsection{Self-Learning}
\cite{Vinyals2015Pointer} proposed Pointer Net, which is trained with supervised learning. For combinatorial optimization problems, however, training a model in this way has some issues: (1) the performance of model depends on the quality of labeled data, (2) getting highly qualified labeled data for learning is not feasible or costly in some combinatorial optimization problems. By contrast, we believe that reinforcement learning, which requires little direction, is a natural framework for learning the value function $h$ in Equation \ref{formula_3}.\\

\noindent \textbf{Reinforcement learning formulation}\\

\noindent We define $states$, $actions$, $rewards$ in the reinforcement learning framework as follows:

\begin{itemize}
	\item $States$: a state $S$ is an ordered sequence of traversed nodes on a graph $G$. We use graph embedding to encode each state as a vector in the $l$-dimensional space. The terminal state $\hat{S}$ is that we have traversed all the nodes.
	\item $Transition$: transition is deterministic in traveling salesman problem, and correspond to adding selected node $v \in \bar{S}$ to $S$, where $\bar{S}$ and $S$ are the traversed sequence and non-traversed sequence respectively.
	\item $Actions$: an action $v$ is a node of $G$ in the non-traversed sequence $\bar{S}$.
	\item $Rewards$: When all nodes in $G$ are traversed, the length $D$ of ordered sequence $\hat{S}=\{v_1,v_2,..,v_n\}$ can be calculated according to following formulation,
	\begin{equation}\label{formula_6}
	f=\sum_{i=1}^{|\hat{S}|-1}w_{i,i+1}+w_{|\hat{S}|,1}
	\end{equation}
	We can also calculate the length of partial sequence $S^{p} = S \cap v$ when the node $v$ is added to $S$ as follows,
	\begin{equation}\label{formula_7}
	g_{v}=\sum_{i=1}^{|S^{p}|-1}w_{i,i+1}
	\end{equation}
	We define the reward function $r(s,v)$ at state $s$ as the length of the partial ordered sequence of $\hat{S}$ where the starting node is $v$. That is,
	\begin{equation}\label{formula_8}
	r(s,v)=f-g_{v}
	\end{equation}
	\item $Policy$: Based on the value function $h$ of neural network, we use Monte Carlo tree search as default policy to select next action $v$. After repeated $t$ times $playout$, we choose a action $v$ among all valid actions of the root state $s$ by following formulation,
	\begin{equation}\label{formula_9}
	v=\mathop{\arg\max}_{v_i\in V_c}\hat{Q}_{v_i}
	\end{equation}
	where $V_c$ is the set of all valid actions of root state $s$, and $Q_{v_i}$ is the reward of state, which is obtained by taking action $v_i$ from the root state.
\end{itemize}

\noindent \textbf{Learning algorithm}\\

\noindent Similar to \cite{Silver2016Mastering}, we perform end-to-end learning of neural network. First, the parameters of neural network are initialized to random weights $\Theta_{0}$. When an episode ends where all the nodes have been traversed, the data for each time-step $t$ is stored as $(s_t,v_t,r_t)$, where $r_t$ can be calculated according to Equation \ref{formula_8}. The neural network is trained from sampling uniformly among all time-steps $(s,v,t)$. Specially, the parameters $\Theta$ are learned by gradient descent on a loss function $l$ over the mean-squared error,
\begin{equation}\label{formula_10}
l=(r-h)^2+c||\Theta||^2
\end{equation}
where c is a parameter that control the level of L2 weight regularization.\\

Our training algorithm, described in Algorithm 1,

	\begin{algorithm}
		\caption{\textbf{Training Algorithm}}
		\begin{algorithmic}[1]
			\State Initialize experience replay memory M to capacity N
			\For{$i = 0 \to MaxEpisode$}
			\State Draw graph G from distribution D
			\State Initialize the state to empty S=()
			\For{$step = 1 \to EndStep$}
				\State {$v_t = \mathop{\arg\max}_{v_i\in \bar{S}}\hat{Q}_{v_i}$}
				\State Add $v_t$ to partial solution: $S_{t+1}:=(S_t,v_t)$
				\EndFor
			\State Add tuple $(S_i,v_i,r_i)$ to M, $i=1,2,...,EndStep$
			\State {Sample random batch from B $\stackrel{\text{i.i.d}}{\sim}$ M}
			\State Update $\Theta$ by Adam over (10) for B
			\EndFor
			\State return $\Theta$
			
		\end{algorithmic}
	\end{algorithm}

\section{Experimental Evaluation}
\subsubsection{Instance generation.}
To evaluate the proposed approach against other deep learning approaches, we generate graph instances by the instance generator from the DIMACS TSP Challenge \cite{johnson2007experimental}. We produce two types of graphs: random instances include $n$ points scattered uniformly at random in the $[10^6, 10^6]$ square and clustered instances includes $n$ points which are clustered into four groups. We use the state-of-the-art solver, Gurobi\footnote{http://www.gurobi.com/} to compute optimal solutions.

\subsubsection{Experimental Details.}
For our approach, the graph representations and hyper-parameters are described as follows. We embed nodes' features to a 64 dimensional vector. We train our method using Adam optimizer \cite{kingma2014adam} and use the learning rate of $10^{-4}$. We use 400 simulations for selecting each move in the Monte Carlo tree search during training and testing. We use Bayesian Optimization to find the best value of the $C_{p}$ and get the best performance when setting $C_{p}$ to 0.5.

\subsubsection{Details on Training and Testing}
We train different models for TSP20 and TSP50 respectively using 40 graphs randomly selected from the dataset. During testing, we use the pre-trained model for TSP20 to evaluate performance on TSP20 and use the pre-trained model for TSP50 to evaluate performance on TSP50. While for TSP100, we use the same model which trained for TSP50. We use 100 graphs to test for the above three problems. Instead of using Active Search in \cite{Bello2016Neural}, we use the pre-trained mode directly to select the best solution among the results which are obtained starting different nodes. 

\subsubsection{Results and Analyses}
We compare our approach with three excellent work, Pointer Network \cite{Vinyals2015Pointer}, S2V-DQN \cite{Dai2017Learning} and AttentionTSP \cite{Kool2018Attention}. We use a machine with CUDA Titan XP for training and testing above three methods. For Pointer network, we do not reproduce successfully the results reported in the paper. We keep the original experimental setup for training S2V-DQN and AttentionTSP. Before we test the new instances generated by us, the performance of S2V-DQN and AttentionTSP has achieved the results as shown in the paper. And then we fine-tune the parameters of the above two methods using data generated by us. Rather than reporting the approximation ration $\frac{c}{c^*}$ we report the average optimality gap $\frac{c-c^*}{c^*}=\frac{c}{c^*}-1$ mentioned in \cite{Kool2018Attention}.\\

We report the average optimality gap of the above approaches on random graphs in Table \ref{tabel:random}. Each approach is trained on random graphs and then tested on random graphs. Our approach performs favorably against Pointer network and gets comparable performances compared with S2V-DQN. 

\begin{table}[H]
	\begin{tabular}{lllc}
		\hline
		Approach      & TSP20      & TSP50      & TSP100  
		\\ \hline 
		Pointer Network          &1.102&	1.128& --	  \\
		AttentionTSP &1.003&	1.017&	1.045 \\
		S2V-DQN          &1.019&	1.062&	1.081 \\
		Our                 & 1.010&	1.063&	1.095 \\\hline
		
		\hline
	\end{tabular}
	\caption{Average optimality gap of different models on random instances. We directly use the result reported in the paper of Pointer Network.}
	\label{tabel:random}
\end{table}

Table \ref{tabel:cluster} is the average optimality gap of the above approaches on clustered graphs. Each approach is trained on random graphs and then tested on clustered graphs. Our approach gets better result than S2V-DQN on TSP20. When the number of nodes in the graph increases from 50 to 100, our approach is more stable than S2V-DQN. What's more, the performance of AttentionTSP is poor on TSP100. Our approach can generalize on different kinds of graphs well than AttentionTSP.

\begin{table}[H]
	\begin{tabular}{lllc}
		\hline
		Approach                    & TSP20    & TSP50    & TSP100   \\ \hline
		Pointer Network          &-- &-- & --	  \\
		S2V-DQN          & 1.027 & 1.061 & 1.082 \\
		AttentionTSP & 1.017 & 1.101 & 1.685 \\
		Our               & 1.025 & 1.106 & 1.109\\\hline
	\end{tabular}
	\caption{Average optimality gap of different models on clustered instances. We exclude Pointer network as the approach do not test on the cluster graphs in the original paper.}
	\label{tabel:cluster}
\end{table}
Besides the experiments for synthetic data, we evaluate our approach on the real-world dataset called TSPLIB \footnote{https://www.iwr.uni-heidelberg.de/groups/comopt/software/\\TSPLIB95/tsp/s}. Due to the limitation of computing resources, we only test the instances whose node's number is less than 100. Our approach can get the comparable performance of S2V-DQN.
\begin{table}[H]
	\begin{tabular}{lllc}
		\hline
		Instance                    & OPT    & Our    & S2V-DQN   \\ \hline
		eil51          &426 &442 & 439	  \\
		berlin52          & 7542 & 7598 &7542  \\
		st70 & 675 & 695 &  696\\
		eil76 & 538 & 545 & 564 \\
		pr76               & 108159 & 108576 & 108446\\
		average optimality gap & 1 & 1.003 & 1.002 \\\hline
	\end{tabular}
	\caption{Best solutions of different models on real-world instances. We also evaluate AttentionTSP on those instances, but its performance is very poor. For example, the best solutions for eil51 and berlin52 are 1733 and 28233 respectively.}
	\label{tabel:cluster}
\end{table}
\section{Conclusion}
We proposed a new framework to solve traveling salesman problem, which combines Monte Carlo tree search and deep reinforcement learning. Inconsistent with previous works in which labeled data or hand-crafted features may occupy an important place, our framework is completely unsupervised and can learn with samples generated by itself. The core idea of our approach lies in converting TSP into tree search problem. Our framework is, to our best of knowledge, the first tree-search combined with the deep neural network method in combinatorial optimization. We have demonstrated that the proposed framework performs favorably against other methods in small-to-medium problem settings. And it shows comparable performance as state-of-the-art in large problem setting.

\bibliographystyle{aaai}
\bibliography{aaai2019_conference.bib}

\end{document}